\documentclass{article}
\pdfoutput=1

\usepackage[preprint,nonatbib]{neurips_2024}

\usepackage[utf8]{inputenc} 
\usepackage[T1]{fontenc}    
\usepackage{hyperref}       
\usepackage{url}            
\usepackage{booktabs}       
\usepackage{amsfonts}       
\usepackage{nicefrac}       
\usepackage{microtype}      
\usepackage{xcolor}         

\usepackage{multirow}
\usepackage{graphicx}
\usepackage{colortbl}
\definecolor{mygray}{gray}{0.95}
\usepackage{pifont}
\newcommand{\cmark}{\ding{51}}%

\title{One Token to Seg Them All: Language Instructed Reasoning Segmentation in Videos}

\author{
    Zechen Bai$^{1}$ \quad
    Tong He$^{2}$ \quad
    Haiyang Mei$^{1}$ \quad
    Pichao Wang$^{2}$ \quad
    Ziteng Gao$^{1}$ \\
\textbf{
    Joya Chen$^{1}$\quad
    Lei Liu$^{2}$\quad
    Zheng Zhang$^{2}$\quad
    Mike Zheng Shou$^{1}$\thanks{Corresponding Author}
} \\
    $^1$Show Lab, National University of Singapore \quad
    $^2$Amazon \\
}

\begin{document}

\maketitle

\begin{abstract}

We introduce VideoLISA, a video-based multimodal large language model designed to tackle the problem of language-instructed reasoning segmentation in videos.
Leveraging the reasoning capabilities and world knowledge of large language models, and augmented by the Segment Anything Model, VideoLISA generates temporally consistent segmentation masks in videos based on language instructions.
Existing image-based methods, such as LISA, struggle with video tasks due to the additional temporal dimension, which requires temporal dynamic understanding and consistent segmentation across frames.
VideoLISA addresses these challenges by integrating a Sparse Dense Sampling strategy into the video-LLM, which balances temporal context and spatial detail within computational constraints.
Additionally, we propose a One-Token-Seg-All approach using a specially designed \texttt{<TRK>} token, enabling the model to segment and track objects across multiple frames.
Extensive evaluations on diverse benchmarks, including our newly introduced ReasonVOS benchmark, demonstrate VideoLISA's superior performance in video object segmentation tasks involving complex reasoning, temporal understanding, and object tracking.
While optimized for videos, VideoLISA also shows promising generalization to image segmentation, revealing its potential as a unified foundation model for language-instructed object segmentation.
Code and model will be available at: \textcolor{purple}{\url{https://github.com/showlab/VideoLISA}}.
\end{abstract}

\section{Introduction}

We live in a dynamic world.
Localizing objects of interest in videos according to human intent is a crucial task for intelligent models and systems.
Language, as a natural interface, serves as the primary reference for identifying target objects.
However, language expressions vary widely across different scenarios, presenting varying levels of difficulty.
While category names are straightforward references, detailed text descriptions from tasks like referring segmentation~\cite{refcoco,refcocog,seo2020urvos} introduce greater complexity.
In real-world applications, these expressions can be more complex, involving intent understanding, reasoning, and world knowledge, making them more user-friendly yet significantly more challenging for models to understand and act upon.

Recent advancements in the image domain have shown progress in language-instructed reasoning for detection and segmentation tasks.
Models leveraging multimodal large language models (MLLMs), such as those in DetGPT \cite{pi2023detgpt} and LISA \cite{lai2023lisa}, have demonstrated the ability to localize target objects by harnessing the implicit reasoning capabilities and world knowledge embedded in large language models (LLMs).
However, these advancements have not seamlessly translated to video tasks, particularly video object segmentation (VOS).
The primary challenge in VOS stems from the additional temporal dimension, which introduces complexities absent in static images.
VOS requires models to 1) on the input side, capture and comprehend the temporal dynamics present in the video; and 2) on the output side, predict temporally consistent segmentation masks across frames.
These challenges render existing image-based methods inadequate for handling video tasks.

In this work, we introduce VideoLISA, a video-based MLLM designed to address language-instructed reasoning segmentation in videos.
Our goal is to segment target objects throughout the entire video based on diverse language queries that necessitate scene understanding, temporal comprehension, and implicit reasoning.
Drawing inspiration from previous works~\cite{pi2023detgpt,lai2023lisa}, we employ an LLM to inherit its complex reasoning capabilities and adopt the Segment Anything Model (SAM) \cite{sam} to produce segmentation masks.
To overcome the unique challenges presented by video data, we propose two key innovations: a Sparse Dense Sampling strategy and a One-Token-Seg-All approach.

To equip the model with video temporal understanding ability, it is necessary to involve multiple frames.
Processing visual features from all sampled frames in full feature resolution is computationally prohibitive due to the large number of tokens.
In pursuit of efficiency, reducing the frame number would limit the perception of temporal dynamics while down-sampling frame features would lose visual details that are essential for dense prediction tasks exemplified by segmentation.
Our intuition is that adjacent frames in videos usually share similar visual contents and features.
Therefore, we leverage this inherent \textit{temporal redundancy} in videos and propose the Sparse Dense Sampling strategy.
It uniformly samples a set of dense frames, preserving full-resolution features (\textit{dense} tokens), and down-samples the remaining interleaved frames to lower resolution (\textit{sparse} tokens).
Dense tokens provide detailed visual information needed for accurate segmentation, while sparse tokens capture the temporal context, ensuring that the model remains aware of motion and changes over time.
This balance allows the model to construct a coherent spatiotemporal narrative without excessive computational demands.

For achieving temporal consistency in segmentation, instead of handling separate representations for each frame, we propose a One-Token-Seg-All approach.
Prior arts~\cite{slot_naming, obj-tracking} reveal that one compact representation can potentially associate the same object across video frames.
In this work, we design a special \texttt{<TRK>} token to segment and track target objects across multiple frames.
Specifically, we incorporate the \texttt{<TRK>} token into the model's vocabulary and utilize its last hidden embedding in the LLM to prompt the mask decoder to produce segmentation masks.
We improve the temporal consistency from two aspects.
First, when generating the \texttt{<TRK>} token, the model `sees' the video content through the temporal module, which serves as the information foundation for cross-frame association.
In addition, during training, the \texttt{<TRK>} token is intentionally trained to segment multiple frames simultaneously, preventing the model from learning shortcuts that focus only on spatial information of a certain frame.
During inference, a single \texttt{<TRK>} token can segment and track objects across an entire video.
The \texttt{<TRK>} token acts as a unified spatiotemporal representation, encapsulating object information across multiple frames and reducing the complexity of handling multiple prompts.

We evaluate our model on a comprehensive range of public benchmarks, including standard video/image referring segmentation, motion-guided video segmentation, and image reasoning segmentation.
To further assess the model's capabilities in complex reasoning, temporal understanding, and object tracking, we introduce the ReasonVOS benchmark.
Extensive experiments and ablation studies demonstrate the effectiveness of our approach.
Although our model is particularly designed for videos, experiments show that it generalizes well on images, making it a potential foundation model for unified language instructed object segmentation. Our contributions:
\begin{itemize}
\item Sparse Dense Sampling Strategy: We devise a sampling strategy for video-LLM training that achieves a balance between temporal context length and spatial visual detail under computational constraints. This strategyis shown to be effective for spatiotemporal dense prediction tasks, exemplified by video object segmentation.

\item One-Token-Seg-All Approach: We design an effective approach for temporal consistent object segmentation in videos by utilizing a special \texttt{<TRK>} token. This strategy demonstrates robust performance in video object segmentation, leveraging the video-LLM learning module and a specially designed training objective.

\item VideoLISA Model: We propose VideoLISA, a video-LLM that democratizes reasoning segmentation to videos. Additionally, we introduce the ReasonVOS benchmark, focusing on complex reasoning, temporal understanding, and object movements. This benchmark, along with a range of public benchmarks, comprehensively validates our model's performance.

\end{itemize}

\section{Related Work}

\subsection{Video Object Segmentation}
In computer vision, video object segmentation is a well-studied task~\cite{vos_survey}.
Specifically, referring video object segmentation (RVOS) aims to segment the target object mentioned in a natural language expression in a video~\cite{seo2020urvos,mttr,wu2023onlinerefer,ye2019cross,li2023robust,wu2022referformer,miao2023spectrum}.
Compared with image segmentation, RVOS is more challenging since both the action and appearance of the referred object must be segmented in a video.
Gavrilyuk et al. (2018) were the first to propose the RVOS task and the A2D-Sentences benchmark~\cite{gavrilyuk2018actor}.
This field continues to evolve with new benchmarks emerge such as Ref-DAVIS-17~\cite{refdavis}, Ref-YouTube-VOS~\cite{seo2020urvos}, and MeViS~\cite{ding2023mevis}.
Many previous studies have primarily adapted referring image segmentation approaches for frame-by-frame object segmentation.
For example, URVOS~\cite{seo2020urvos} and RefVOS~\cite{refvos} utilize cross-modal attention for per-frame segmentation.
Some recent works, such as ReferFormer~\cite{wu2022referformer} and MTTR~\cite{mttr}, employ a DETR-like structure, which simplifies the referring pipeline and achieves impressive performance.
R2VOS~\cite{li2023robust} enhances multi-modal alignment through text reconstruction. 
OnlineRefer~\cite{wu2023onlinerefer} proposes an online model with explicit query propagation.
SgMg~\cite{miao2023spectrum} proposes a segment-and-optimize paradigm to solve the feature drift issue.
Despite the impressive results achieved by these methods, several challenges remain.
First, most existing methods are deficient in comprehending the motion information in videos and languages, as revealed by the recent MeViS~\cite{ding2023mevis} benchmark.
Second, there are few studies on complex reasoning-based segmentation in the video domain, both methodologically and benchmark-wise.

\subsection{Multimodal Large Language Model}
\label{sec:related_work_mllm}

The remarkable advancements of large language models (LLMs) motivate the research community to extend the foundational capabilities of LLMs to the visual domain, leading to multimodal large language models (MLLMs)~\cite{mllm_survey,mllm_hallu_survey}.
The pioneering works of MLLMs, such as LLaVA~\cite{llava}, MiniGPT-4~\cite{zhu2023minigpt}, and InstructBLIP~\cite{dai2024instructblip}, exhibit impressive visual understanding capabilities, including image captioning~\cite{recall_cap,bai2021explain} and visual question answering.
When extending into the video domain, a prominent issue is handling the temporal dimension.
One straightforward approach is to concatenate the tokens from multiple frames~\cite{video-llava}, though the temporal length might be limited by computational resources.
To address this, one line of work~\cite{video-chat-gpt,chatuniv,huang2024lita} explores pooling (merging) strategies to reduce the number of tokens, such as pooling along the spatial and temporal dimensions separately~\cite{video-chat-gpt}, token merging based on similarity~\cite{chatuniv}, and pooling with different strengths at a slow-fast pace~\cite{huang2024lita}.
Another line of work~\cite{li2023videochat,videollama} utilizes the Q-former~\cite{blip2} architecture to extract abstracted features, which greatly reduces the number of tokens.

More recently, some studies have further integrated region-level image understanding and grounding abilities into MLLMs.
Kosmos-2~\cite{peng2023kosmos} and Shikra~\cite{chen2023shikra} directly quantize bounding boxes into discrete location tokens or numeric representations of positions.
GPT4RoI~\cite{zhang2023gpt4roi} uses a simple pooling operation to extract features within boxes or masks as the region representations.
Another line of work leverages the reasoning ability of MLLMs and resorts to off-the-shelf models for localization.
For example, DetGPT~\cite{pi2023detgpt} utilizes a pre-trained LLM and an open-vocabulary object detector to detect the target object based on human intent described in natural language.
LISA~\cite{lai2023lisa} connects an MLLM and the Segment Anything (SAM)~\cite{sam} model using a special token to produce fine-grained segmentation masks.
Although these works have achieved impressive performance on image tasks, they are still incapable of processing videos.
For object segmentation in videos, very few studies have leveraged the reasoning ability of LLMs to overcome current limitations.
PG-Video-LLaVA~\cite{pg-video-llava} utilizes off-the-shelf object detector and tracker to obtain the target objects first and then match it with the entities mentioned in the generated text.
TrackGPT~\cite{trackgpt} makes a straightforward extension of LISA by iteratively updating the special token with video progresses.
However, the absence of video learning module significantly limits its perception and reasoning of temporal dynamics.

\section{Method}

\begin{figure*}[ht]
    \centering
    \includegraphics[width=0.95\linewidth]{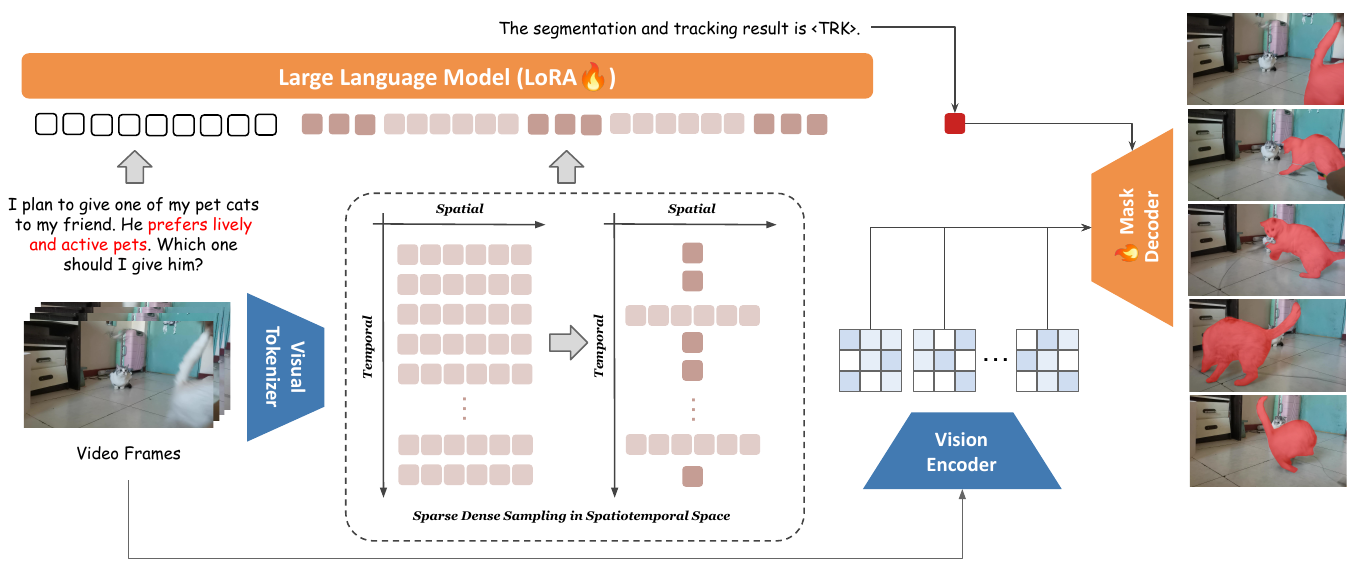}
    \vspace{-6pt}
    \caption{Framework of our approach.
    }
    \vspace{-12pt}
    \label{fig:method}
\end{figure*}

The task of language-instructed reasoning segmentation in videos can be formally defined as follows.
Given a video $\mathcal{X}_{\rm vid}$ and a language expression $\mathcal{X}_{\rm txt}$, the model takes both as input and outputs the pixel-level segmentation masks $\mathcal{M}$ for all frames.
$\mathcal{X}_{\rm txt}$ is a free-form text that particularly emphasizes implicit intent reasoning, world knowledge, and video temporal dynamics.

\subsection{Architecture}

Fig.~\ref{fig:method} illustrates the model architecture.
It consists of a visual tokenizer, an LLM, a vision encoder, and a promptable mask decoder.
We omit the text tokenizer in the LLM for simplicity.
The visual tokenizer and LLM are initialized from LLaVA~\cite{llava,llava-pp}.
The vision encoder and mask decoder are initialized from SAM~\cite{sam}.
Given a video, we first uniformly sample $T_{\rm sparse}$ frames and encode them into visual tokens via the visual tokenizer, resulting in $T_{\rm sparse}\times L$ tokens in total.
Ideally, larger $T_{\rm sparse}$ would be better for capturing temporal dynamics.
However, it is prohibitive to let the LLM process such a large number of tokens.
Thus, we develop the Sparse Dense Sampling strategy to reduce the number of tokens, which will be elaborated in Sec.~\ref{sec:sparse-dense-sampling}.
After that, the visual tokens are concatenated with text tokens and fed into the LLM.

To equip the LLM with segmentation capabilities, following previous work~\cite{lai2023lisa}, we extend the vocabulary of the LLM with a special token \texttt{<TRK>}.
During generation, this special token carries rich semantic information from the text prompt and video content, providing signals for decoding pixel-level segmentation masks.
Specifically, we extract the last layer embedding corresponding to the \texttt{<TRK>} token and transform it into a prompt embedding with a multi-layer perceptron (MLP).
At the same time, the vision encoder extracts per-frame features from the video.
Finally, the prompt embedding and the visual features are processed by the mask decoder to produce the segmentation masks.
Note that for one video, there is only one prompt embedding that is in charge of all the frames.
The One-Token-Seg-All approach will be introduced in Sec.~\ref{sec:one-token-seg-all}.

\subsection{Sparse Dense Sampling}
\label{sec:sparse-dense-sampling}

Given the $T_{\rm sparse}\times L$ tokens, we aim to reduce the number of tokens while preserving enough spatial details and temporal dynamics.
Therefore, we further sample $T_{\rm dense}$ frames out of $T_{\rm sparse}$ frames.
The visual tokens of the $T_{\rm dense}$ frames are all preserved in full resolution, \textit{i.e., dense} tokens. 
Then, we apply global average pooling on the $T_{\rm sparse}$ frames to reduce them to low resolution, \textit{i.e., sparse} tokens. 
In our implementation, each frame is represented by only one token.
Finally, the total number of tokens is reduced to $T_{\rm sparse} + T_{\rm dense} \times L$, which is significantly smaller than $T_{\rm sparse} \times L$.
The rationale behind this strategy is the inherent temporal redundancy in video data.
By exploiting this, we reduce the computational burden without losing critical information.
The dense tokens provide visual details for their adjacent sparse frames, while the sparse tokens capture the temporal dynamics for the dense frames.
In Sec.~\ref{sec:related_work_mllm}, we have discussed several popular temporal learning strategies in video-LLM.
Although they exhibit remarkable performance in general video understanding tasks, our empirical studies (see Tab.~\ref{tab:abl_temporal_model_mini}) demonstrate that these popular strategies are not seamlessly transferable to video object segmentation.
This is likely because they either lose spatial details or temporal information, both of which are essential in dense prediction tasks in videos.

\vspace{-3pt}
\subsection{One Token Seg All}
\vspace{-3pt}
\label{sec:one-token-seg-all}

\begin{figure*}[ht]
    \centering
    \includegraphics[width=0.85\linewidth]{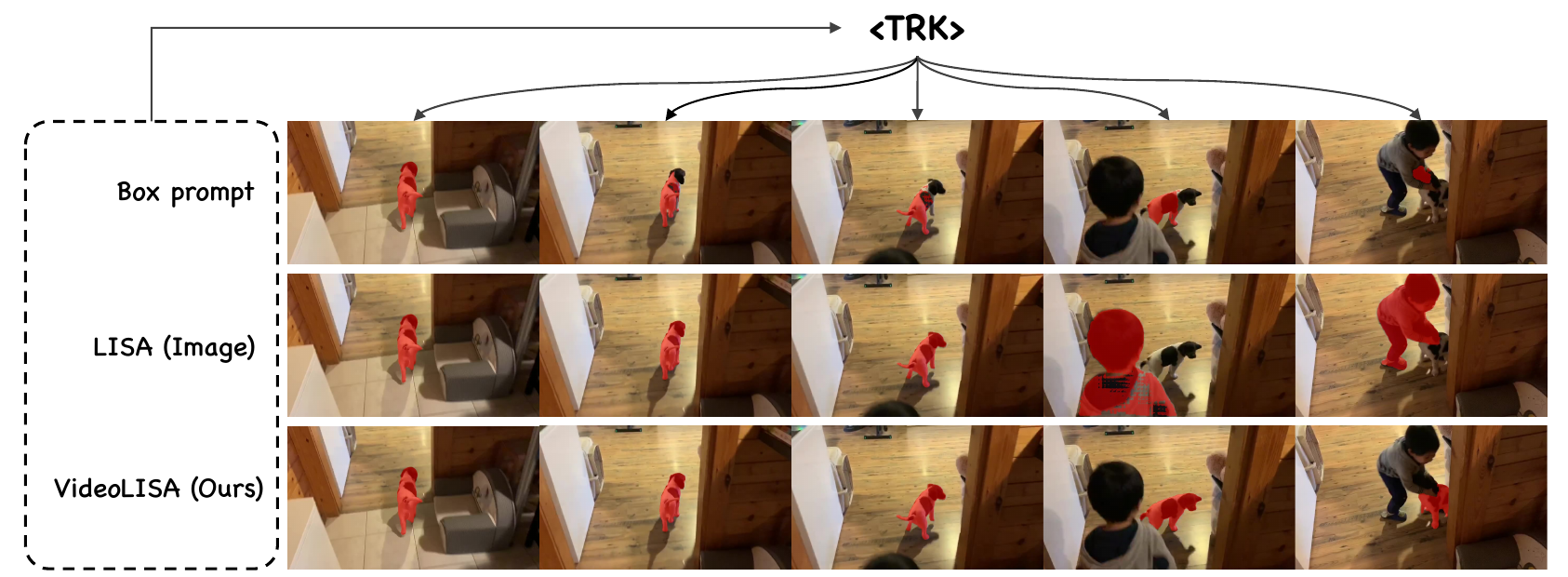}
    \vspace{-6pt}
    \caption{Exploration of One-Token-Seg-All approach.
    }
    \vspace{-12pt}
    \label{fig:one-token-seg-all}
\end{figure*}

As shown in Fig.~\ref{fig:method}, throughout the video, we use a single special \texttt{<TRK>} token for segmenting all the frames.
We provide an in-depth analysis of the rationale behind this approach.
In our model, the promptable segmentation model is initialized from SAM, in which the decoder takes the prompt embedding and visual features as inputs and outputs masks.
Our intuition is that \textit{segmenting one object in multiple frames can be regarded as segmenting multiple regions (instances) in one image grid}.
From this perspective, SAM~\cite{sam} itself already has the potential to segment objects across multiple frames, \textit{if the prompt is properly given}.
Previous works~\cite{slot_naming, obj-tracking, fan2024adaptive} suggest that one compact representation has the potential to associate the same entity across video frames.
For example, from the perspective of object tracking~\cite{obj-tracking}, the prompt embedding can be regarded as a semantic kernel while the visual features are the context to be contrasted.
This motivates us to explore whether one prompt embedding is capable of tracking under the promptable decoding paradigm of SAM.

To answer this question, one key problem is whether the prompt embedding contains enough semantic information to serve as the kernel.
In SAM, its own prompt encoder mainly accepts visual prompts, such as points, boxes, and masks.
In videos, the object moves dynamically.
Our pilot study in Fig.~\ref{fig:one-token-seg-all} shows that visual prompts quickly fail in the presence of object motion.
This is expected since these visual prompts heavily rely on the object's spatial location.
We then explore the prompt embedding produced by an image reasoning segmentation model, LISA~\cite{lai2023lisa}, which employs a LLM and is trained with segmentation data.
It can be expected that its prompt embedding should contain more semantic information, at least significantly more than that of the visually instructed prompt.
The second row of Fig.~\ref{fig:one-token-seg-all} validates this hypothesis by applying one prompt embedding to multiple frames.
Compared to box prompts, the prompt embedding from LISA shows improved resilience to object movement, as demonstrated in the first three frames.
However, when the object's motion becomes larger and a distractor object appears, the segmentation fails again, drifting to another object nearby.

We identify two primary factors that account for the failure.
Firstly, the input of LISA model only has one frame, which contains very limited temporal information.
Therefore, the generated prompt embedding lacks the information required for cross-frame association.
Secondly, during the training of LISA, the prompt embedding is trained to segment only one frame.
This potentially allows it to learn a shortcut that merely encompasses positional information, rather than learning the semantic information that generalizes across frames.
In our work, the approach of using one token to segment multiple frames has been developed by addressing these issues accordingly.
Firstly, the Sparse Dense Sampling-based temporal learning module provides spatiotemporal information of the video.
The model `sees' the video content, which is the foundation of mask association.
Furthermore, during training, we intentionally train the \texttt{<TRK>} token to segment multiple frames.
This objective would enforce the token to learn more `semantic' information that can be used as the semantic kernel and segment the target object across frames.
The last row of Fig.~\ref{fig:one-token-seg-all} presents the segmentation and tracking produced by the \texttt{<TRK>} token in our VideoLISA.

\vspace{-5pt}
\subsection{Training and Inference}
\noindent\textbf{Training Data.}
The training data for our model mainly consists of two parts: 1) image segmentation and 2) video segmentation.
For the image part, we follow the setting of LISA~\cite{lai2023lisa}.
For the video data, we employ video object segmentation (VOS) and referring video segmentation data (RVOS).
During pre-processing, we fill the original category name or referring expression in the dataset into a template.
For example: ``\texttt{\textbf{USER}: <VIDEO> Can you segment \{description\} in this scene? \textbf{ASSISTANT}: Sure, it is <TRK>.}'', where \{\texttt{description}\} is the placeholder to fill.
For VOS data that contain videos with multi-class labels, we randomly choose one class and merge all the masks belonging to this class into one binary mask.
\noindent\textbf{Training Objective.}
The model is trained end-to-end using the text generation loss $\mathcal{L}_{\rm txt}$ and segmentation loss $\mathcal{L}_{\rm seg}$.
The segmentation loss consists of per-pixel binary cross-entropy (BCE) loss and DICE loss.
The final loss is computed as the weighted sum of the three losses.
For video training, we compute the segmentation loss on the sampled $T_{\rm dense}$ frames in parallel and average them.

\noindent\textbf{Inference.}
During inference, given a video, $T_{\rm sparse}$ and $T_{\rm dense}$ frames are sampled similarly to training, except that the $T_{\rm dense}$ frames are uniformly sampled from $T_{\rm sparse}$ rather than randomly.
After obtaining the \texttt{<TRK>} token from the LLM, we feed all the frames of the video into the mask decoder one by one, using the same \texttt{<TRK>} token to segment each frame, yielding a list of masks.
\noindent\textbf{Post optimization.}
Among these frames, the $T_{\rm dense}$ frames are seen in full resolution by the model, making their segmentation masks more reliable and accurate.
For the remaining frames, although the One-Token-Seg-All strategy exhibits impressive cross-frame segmentation performance, our empirical observations indicate it inevitably suffers from low mask quality, likely limited by the inherent capability of the SAM model.
Thus, we employ post-optimization as an optional step to further enhance mask quality.
Specifically, we take XMem++~\cite{bekuzarov2023xmem++} as the post-optimization approach.
Compared to XMem~\cite{cheng2022xmem}, which propagates one mask through the video, XMem++ distinguishes itself by taking multiple `reliable' masks as reference and inferring the masks of the remaining frames.
This paradigm is naturally suitable for our method since the $T_{\rm dense}$ frames span uniformly across the video, providing long-range yet diverse masks as references.

\vspace{-7pt}
\section{Benchmark}
\vspace{-8pt}
\label{sec:benchmark}

The versatile abilities of our model can be evaluated using public benchmarks that assess various aspects.
RVOS benchmarks~\cite{refdavis,seo2020urvos} evaluate temporal-related abilities, involving referring expression comprehension, video temporal understanding, and temporal consistent segmentation.
Complex reasoning abilities can be assessed by the image-based reasoning segmentation benchmark~\cite{lai2023lisa}.
However, there is still a lack of a benchmark that comprehensively evaluates the reasoning segmentation abilities of videos.
Towards this goal, we have organized the \textit{ReasonVOS} benchmark.
Specifically, we annotate language expressions based on the videos and mask annotations from existing datasets, including MOSE~\cite{ding2023mose}, MeViS~\cite{ding2023mevis}, VIPSeg~\cite{vipseg}, and BURST~\cite{athar2023burst}.
The criteria for data collection and annotation processes are as follows.
Each language expression should encompass at least one of the following aspects: 1) complex reasoning, 2) world knowledge, 3) temporal dynamics.
For the video and mask selection, objects with explicit movement are highly prioritized to evaluate the temporal consistency of masks.
As a result, we manually annotated 105 samples as initial seed data.
Following previous practices~\cite{ding2023mevis,trackgpt}, we use a LLM to rephrase the language expressions for augmentation and perform another round of human checking.
The resulting ReasonVOS benchmark comprises 458 video-instruction-mask data samples.
This benchmark is specifically designed for zero-shot evaluation purposes, as the reasoning ability is embedded in the LLM and can be triggered by existing image-based reasoning segmentation data.

\vspace{-8pt}
\section{Experiments}
\vspace{-5pt}

\subsection{Experimental Setting}
\vspace{-5pt}
\label{sec:exp_setup}

\noindent\textbf{Datasets}
Our model is trained on a variety of segmentation datasets.
The image-based datasets include
1) semantic segmentation: ADE20K~\cite{ade20k}, COCO-Stuff~\cite{cocostuff}, PACO-LVIS~\cite{ramanathan2023paco}, and PASCAL-Part~\cite{pascal_part};
2) referring segmentation: refCLEF, refCOCO, refCOCO+~\cite{refcoco}, and refCOCOg~\cite{refcocog};
3) reason segmentation: 239 ReasonSeg samples from LISA~\cite{lai2023lisa}.
The video-based datasets we use include:
1) semantic VOS: YouTube-VOS~\cite{vos2019};
2) referring VOS: Refer-YouTube-VOS~\cite{seo2020urvos} and MeViS~\cite{ding2023mevis}.
The evaluation benchmarks will be elaborated in the corresponding experiment sections.

\noindent\textbf{Implementation Details}
We implement our model with LLaVA-Phi-3-V~\cite{llava-pp}, a multimodal LLM based on Phi-3~\cite{phi3} with 3.8B parameters.
We adopt the vision encoder and mask decoder from SAM~\cite{sam}.
We conduct joint training using both image and video datasets.
For video data, we set $T_{\rm sparse}=32$ and $T_{\rm dense}=4$ according to our GPU memory.
For image data, we duplicate the images as pseudo video data.
We train our model using 64 NVIDIA 24G A10 GPUs with a distributed training script based on DeepSpeed~\cite{rasley2020deepspeed}.
We use the AdamW~\cite{adamw} optimizer with the learning rate and weight decay set to 0.0003 and 0, respectively.
We also adopt WarmupDecayLR as the learning rate scheduler, with the warmup iterations set to 100.
The weights of the text generation loss ($\lambda_{\rm txt}$) and the mask loss ($\lambda_{\rm seg}$) are both set to 1.0.
The weights of the BCE loss ($\lambda_{\rm bce}$) and the DICE loss ($\lambda_{\rm dice}$) are set to 2.0 and 0.5, respectively.
The per-device batch size is set to 2.
For ablation studies, the total number of iterations is $3,000$ and each experiment takes around 10 hours.
For the final model used for comparison, we scale up the training to $6,000$ iterations, which takes 20 hours.

\noindent\textbf{Evaluation Metrics}
For image-based evaluation, we adopt two metrics commonly used in previous works~\cite{refcoco,lai2023lisa}: gIoU and cIoU.
gIoU is defined by the average of all per-image Intersection-over-Unions (IoUs), while cIoU is defined by the cumulative intersection over the cumulative union.
For video-based evaluation, we follow previous practices~\cite{wu2022referformer,wu2023onlinerefer} and use region similarity (J), contour accuracy (F), and their average value (J\&F).

\subsection{Evaluation on Video Tasks}

\vspace{-3pt}
\subsubsection{Referring Video Object Segmentation}
\vspace{-3pt}
We adopt two benchmarks of standard referring video object segmentation.
Ref-Youtube-VOS is evaluated on the official challenge server~\footnote{\url{https://codalab.lisn.upsaclay.fr/competitions/3282}}.
Ref-DAVIS-17 is evaluated by the official evaluation code~\footnote{\url{https://github.com/davisvideochallenge/davis2017-evaluation}}.
The evaluation results are shown in Tab.~\ref{table:sota_refer}.
Our method demonstrates competitive performance on both benchmarks, achieving comparable or superior results to existing methods.
For Refer-DAVIS-17, our method achieves state-of-the-art performance, outperforming all the other methods by a considerable margin.
In Refer-YouTube-VOS, our method performs well compared to traditional RVOS methods, achieving a high rank.
State-of-the-art methods, such as SgMg~\cite{miao2023spectrum}, achieve remarkable performance, thanks to its dedicated video backbones, such as Video-Swin~\cite{video-swin}.
However, among LLM-based methods with reasoning ability, our model, despite having only 3.8B parameters, outperforms other methods with much larger LLMs, such as LISA-13B and TrackGPT-13B.

\begin{table*}[t]
	\centering
	\small
	\caption{The quantitative evaluation results on Refer-Youtube-VOS and Refer-DAVIS-17.
    In the table, \textbf{bold} denotes the best scores; \underline{underline} denotes the second place.
 }
	\resizebox{0.80\textwidth}{!}{
		\setlength\tabcolsep{8pt}
		\renewcommand\arraystretch{1.0}
		\begin{tabular}{l|c|ccc|ccc}
			\specialrule{.1em}{.05em}{.05em}
			\multirow{2}{*}{Method} & \multirow{2}{*}{Year} & \multicolumn{3}{c|}{Refer-Youtube-VOS} &\multicolumn{3}{c}{Refer-DAVIS-17}  \\
			\cline{3-8}
			
			& &$\mathcal{J}$\&$\mathcal{F}$& $\mathcal{J}$ & $\mathcal{F}$& $\mathcal{J}$\&$\mathcal{F}$ & $\mathcal{J}$ & $\mathcal{F}$ \\ 
			\hline
            \rowcolor{mygray}
            \multicolumn{8}{l}{\emph{Traditional methods without reasoning ability}} \\
            \hline
            URVOS~\cite{seo2020urvos}&2020 & 47.2 & 45.2 & 49.1  & 51.6&  47.2 & 55.9 \\
			CMPC-V~\cite{liu2021cross} &2021 &47.5 &45.6 &49.3 &-&-&- \\
            YOFO~\cite{li2022you} &2022 &48.6 &47.5 &49.7 &53.3 &48.8 &57.8 \\
			LBDT~\cite{Ding_2022_CVPR} &2022 &49.4 &48.2 &50.6 & 54.3 & -&- \\
            MLSA~\cite{wu2022multi} &2022 &49.7 & 48.4 & 50.9 & 57.9 & 53.8 & 62.0 \\
			PMINet + CFBI~\cite{ding2021progressive}  &2021  & 54.2 & 53.0 & 55.5 & - & - & - \\
            MTTR~\cite{mttr}  &2022 & 55.3 & 54.0 & 56.6 & - & - & - \\
			CITD~\cite{liang2021rethinking}  &2021 & 61.4 & 60.0 & 62.7  & - & - & -\\
			ReferFormer~\cite{wu2022referformer}  &2022 & 62.9 & {61.3} & 64.6 & 61.1 &58.1 &64.1\\
            R$^2$-VOS~\cite{li2023robust} &2023 &61.3 &59.6 &63.1 &-&-&- \\ 
			SgMg~\cite{miao2023spectrum} &2023 &\textbf{65.7}&\textbf{63.9} &\textbf{67.4}&63.3&60.6 &66.0 \\
			OnlineRefer~\cite{wu2023onlinerefer} &2023 & 63.5 & 61.6 & 65.5 & 64.8 & 61.6 & 67.7 	\\
            \hline
            \rowcolor{mygray}
            \multicolumn{8}{l}{\emph{LLM-based methods with reasoning ability}} \\
            \hline
			LISA-7B~\cite{lai2023lisa} &2023 &50.2&49.7&50.6&58.4&54.9&61.9\\
			LISA-13B~\cite{lai2023lisa} &2023 &52.6&52.1&53.0&60.7&56.8&64.6\\
			TrackGPT-7B~\cite{trackgpt}  &2023  &56.4  &55.3 &57.4 &63.2  &59.4  &67.0 \\
			TrackGPT-13B~\cite{trackgpt} &2023 &59.5  &58.1  &60.8 & 66.5 & 62.7 & 70.4\\
            
            \textbf{VideoLISA-3.8B (One-Token-Seg-All)} & 2024 & 61.7 & 60.2 & 63.3 & \underline{67.7} & \underline{63.8} & \underline{71.5} \\
            \textbf{VideoLISA-3.8B (Post-optimization)} & 2024 & \underline{63.7} & \underline{61.7} & \underline{65.7} & \textbf{68.8} & \textbf{64.9} & \textbf{72.7} \\

            \specialrule{.1em}{.05em}{.05em}
	\end{tabular} }
	\vspace{-8pt}
	\label{table:sota_refer}
\end{table*}
\begin{table}[t!]
    \begin{minipage}[t]{0.48\textwidth}
        \centering
        \caption{Results on MeViS benchmark.}
           \scalebox{0.7}{\setlength{\tabcolsep}{0.6mm}{\begin{tabular}{l|l|ccc}
                 \specialrule{.1em}{.05em}{.05em}
                 Methods& Year&$\mathcal{J\&F}$ & $\mathcal{J}$ & $\mathcal{F}$ \\
                 \hline
                 URVOS~\cite{seo2020urvos} & 2020 &27.8&25.7&29.9\\
                 LBDT~\cite{Ding_2022_CVPR} & 2022 &29.3&27.8&30.8\\
                 MTTR~\cite{mttr} & 2022 & 30.0&28.8&31.2\\
                 ReferFormer~\cite{wu2022referformer} & 2022 &31.0&29.8&32.2\\
                 VLT+TC~\cite{vltpami} & 2021 &35.5&33.6&37.3\\
                 LMPM~\cite{ding2023mevis} & 2023 & 37.2 &34.2& 40.2\\
                 \midrule
                 \textbf{VideoLISA-3.8B (One-Token-Seg-All)} & 2024 & 42.3 & 39.4 & 45.2 \\
                 \textbf{VideoLISA-3.8B (Post-optimization)} & 2024 & \textbf{44.4} & \textbf{41.3} & \textbf{47.6} \\
                 \specialrule{.1em}{.05em}{.05em}
              \end{tabular}}}
           \label{tab:MeViS}
    \end{minipage}
    \hfill
    \begin{minipage}[t]{0.48\textwidth}
        \centering
        \caption{
        Results on ReasonVOS benchmark.
        }
           \scalebox{0.7}{\setlength{\tabcolsep}{0.6mm}{\begin{tabular}{l|l|ccc}
                 \specialrule{.1em}{.05em}{.05em}
                 Methods& Year&$\mathcal{J\&F}$ & $\mathcal{J}$ & $\mathcal{F}$ \\
                 \hline

                MTTR~\cite{mttr} & 2022 & 31.1 & 29.1 & 33.1 \\
                ReferFormer~\cite{wu2022referformer} & 2022 & 32.9 & 30.2 & 35.6 \\
                SOC~\cite{luo2024soc} & 2023 & 35.9 & 33.3 & 38.5 \\
                 OnlineRefer~\cite{wu2023onlinerefer} & 2023 & 38.7 & 34.6 & 42.9 \\
                 
                 SgMg~\cite{miao2023spectrum} & 2023 & 36.2 & 33.7 & 38.7  \\
                 LISA~\cite{lai2023lisa} & 2023  & 31.1 & 29.1 & 33.1 \\
                 \midrule
                 \textbf{VideoLISA-3.8B (One-Token-Seg-All)} & 2024 & 45.1 & 43.1 & 47.1 \\
                 \textbf{VideoLISA-3.8B (Post-optimization)} & 2024 & \textbf{47.5} & \textbf{45.1} & \textbf{49.9} \\
                 
                 \specialrule{.1em}{.05em}{.05em}
              \end{tabular}}}
        \label{tab:reason_vos} 
    \end{minipage}
    \vspace{-5pt}
\end{table}

\subsubsection{Motion-guided Video Object Segmentation}
We further evaluate our model on motion-guided VOS using the MeViS~\cite{ding2023mevis} benchmark.
Consistent with previous studies~\cite{ding2023mevis,he2024decoupling}, we evaluate our model's performance on the validation set of the MeViS benchmark.
The results in Tab.~\ref{tab:MeViS} demonstrate that our method achieves state-of-the-art performance in this benchmark, outperforming previous methods by a large margin.
We attribute this performance gap to our model's adeptness in capturing temporal dynamics and cross-modal interaction, facilitated by the Sparse Dense Sampling-based temporal module and the One-Token-Seg-All training paradigm.

\subsubsection{Reasoning Video Object Segmentation}
In Tab.~\ref{tab:reason_vos}, we compare various methods on the newly organized ReasonVOS benchmark.
For traditional VOS methods, the metrics are evaluated using their released checkpoints pre-trained on the Ref-YouTube-VOS dataset.
This benchmark focuses on complex reasoning, temporal understanding, and segmentation temporal consistency, which present significant challenges for existing VOS methods and image-based reasoning segmentation methods.
It can be observed that most previous methods exhibit unsatisfactory performance on this benchmark.
Traditional RVOS methods, such as ReferFormer~\cite{wu2022referformer}, excel at tracking moving objects but struggle with comprehending complex language expressions, particularly those requiring multi-step reasoning with world knowledge.
On the other hand, LLM-based models, like LISA~\cite{lai2023lisa}, have better language understanding and reasoning capabilities.
The main reasons for the poor performance are: 1) incapability to capture temporal dynamics in the video, and 2) difficulty in segmenting temporally consistent masks.
In contrast, our VideoLISA model demonstrates remarkable performance, thanks to the advanced model design that considers all these crucial aspects.

\subsection{Evaluation on Image Tasks}
\begin{table}[t]
    \centering
    \caption{Reasoning segmentation results among ours and previous related works. `ft' denotes using 239 reasoning segmentation image-instruction pairs to finetune the model.}
    \vspace{-5pt}
    \tabcolsep=0.2cm
    \resizebox{0.7\textwidth}{!}{{
        \begin{tabular}{ l | c c | c c | c c | c c }
            \toprule
            
            \multirow{3}*{Method} & \multicolumn{2}{c|}{val} & \multicolumn{6}{c}{test} \\ 
            
            \specialrule{0em}{0pt}{1pt}
            \cline{2-9}
            \specialrule{0em}{0pt}{1pt}
            
            ~ & \multicolumn{2}{c|}{overall} & \multicolumn{2}{c|}{short query} & \multicolumn{2}{c|}{long query} & \multicolumn{2}{c}{overall} \\

            \specialrule{0em}{0pt}{1pt}
            \cline{2-9}
            \specialrule{0em}{0pt}{1pt}
            
            ~ & gIoU & cIoU & gIoU & cIoU & gIoU & cIoU & gIoU & cIoU \\ 
            
            \specialrule{0em}{0pt}{1pt}
            \hline
            \specialrule{0em}{0pt}{1pt}

            OVSeg~\cite{liang2023open} & 28.5 & 18.6 & 18.0 & 15.5 & 28.7 & 22.5 & 26.1 & 20.8  \\
            GRES~\cite{liu2023gres} & 22.4 & 19.9 & 17.6 & 15.0 & 22.6 & 23.8 & 21.3 & 22.0 \\    %
            X-Decoder~\cite{zou2023generalized} & 22.6 & 17.9 & 20.4 & 11.6 & 22.2 & 17.5 & 21.7 & 16.3 \\
            SEEM~\cite{zou2023segment} & 25.5 & 21.2 & 20.1 & 11.5 & 25.6 & 20.8 & 24.3 & 18.7 \\
            Grounded-SAM~\cite{groundedsam} & 26.0 & 14.5 & 17.8 & 10.8 & 22.4 & 18.6 & 21.3 & 16.4 \\
            LISA-7B~\cite{lai2023lisa} & 44.4 & 46.0 & 37.6 & 34.4 & 36.6 & 34.7 & 36.8 & 34.1 \\
            LISA-7B (ft)~\cite{lai2023lisa} & 52.9 & 54.0 & 40.6 & 40.6 & 49.4 & 51.0 & 47.3 & 48.4 \\
            LISA-13B~\cite{lai2023lisa} & 48.9 & 46.9 & 39.9 & \textbf{43.3} & 46.4 & 46.5 & 44.8 & 45.8 \\
            LISA-13B (ft)~\cite{lai2023lisa} & 56.2 & 62.9 & \textbf{44.3} & 42.0 & 54.0 & 54.3 & 51.7 & 51.1 \\
            
            \specialrule{0em}{0pt}{1pt}
            \hline
            \specialrule{0em}{0pt}{1pt}

            \textbf{VideoLISA-3.8B (Ours)} & \textbf{61.4} & \textbf{67.1} & 43.8 & 42.7 & \textbf{56.9} & \textbf{57.7} & \textbf{53.8} & \textbf{54.4} \\
            
            \bottomrule            
        \end{tabular}
    }}
    \label{table:reason_seg}   
\vspace{-8pt}
\end{table}
We use the image reasoning segmentation benchmark~\cite{lai2023lisa} to assess the reasoning capability of our model.
During testing, we duplicate an image into multiple frames as a pseudo video.
The results are shown in Tab.~\ref{table:reason_seg}.
We observe that our VideoLISA achieves state-of-the-art performance on both validation set and test set.
Remarkably, despite our model employing an LLM with significantly fewer parameters, it outperforms larger models, such as LISA-7B and LISA-13B, demonstrating its exceptional reasoning capability.
We attribute the impressive performance to the following aspects.
From a data perspective, VideoLISA benefits from joint training on both image and video datasets, allowing it to learn from more abundant and diverse supervision signals.
On the model aspect, the temporal learning module and the One-Token-Seg-All training encourage the model to leverage multiple frames of video simultaneously to conduct reasoning, rather than focusing on one image.
Even when generalizing to image tasks, where the video is simulated by an image, the model's reasoning capability remains effective.
We provide more experiment results on image referring segmentation in the appendix.
These experiments demonstrate that our model is capable of image-based tasks, suggesting the potential for unifying image/video referring/reasoning segmentation tasks into a language-instructed object segmentation task solvable by a single VideoLISA model.

\subsection{Ablation Studies}
\begin{table}[t!]
    \begin{minipage}[t]{0.55\textwidth}
        \centering
        \caption{Ablation study on the temporal modeling architecture.
        }
	       \resizebox{0.95\textwidth}{!}{
		\setlength\tabcolsep{8pt}
		\renewcommand\arraystretch{1.0}
		\begin{tabular}{l|cc|ccc}
			\specialrule{.1em}{.05em}{.05em}
			\multirow{2}{*}{Method} & \multicolumn{2}{c|}{ReasonSeg (val)} &\multicolumn{3}{c}{MeViS (valid\_u)} \\
			\cline{2-6}
			& giou & ciou & $\mathcal{J}$\&$\mathcal{F}$ & $\mathcal{J}$ & $\mathcal{F}$ \\ 
            \midrule
            LISA-7B (Baseline) & 51.7 & 56.7 & 43.2 & 39.9 & 46.5  \\
            LISA-7B (Vid. FT)  & 48.6 & 56.2 & 44.8 & 41.1 & 48.6  \\
            VideoLISA-3.8B ($n$-frame)  & 55.6 & \textbf{60.8} & 49.9  & 46.7 & 53.0 \\
            VideoLISA-3.8B (ST Pooling~\cite{video-chat-gpt})  & 56.0 & 59.9 & 50.8 & 47.8 & 53.8 \\
            VideoLISA-3.8B (Slow-Fast Pooling~\cite{huang2024lita}) & 54.0 & 54.4 & 50.2 & 47.2 & 53.1 \\
            VideoLISA-3.8B (Sparse Dense Sampling) & \textbf{58.9} & 60.0 & \textbf{51.7} & \textbf{48.4} & \textbf{54.9} \\
            \specialrule{.1em}{.05em}{.05em}
	\end{tabular} }
	\label{tab:abl_temporal_model_mini}
    \end{minipage}
    \hfill
    \begin{minipage}[t]{0.42\textwidth}
        \centering
        \caption{Ablation study on the mask association \textit{i.e.,} tracking architecture.
        }
	       \resizebox{0.98\textwidth}{!}{
		\setlength\tabcolsep{8pt}
		\renewcommand\arraystretch{1.0}
		\begin{tabular}{l|ccc}
			\specialrule{.1em}{.05em}{.05em}
			\multirow{2}{*}{Method} &\multicolumn{3}{c}{MeViS (valid\_u)}  \\
			\cline{2-4}
			& $\mathcal{J}$\&$\mathcal{F}$ & $\mathcal{J}$ & $\mathcal{F}$ \\ 
            \midrule
            LISA-7B (Baseline) & 43.2 & 39.9 & 46.5 \\
            LISA-7B + XMem\cite{cheng2022xmem} & 45.6 & 41.9 & 49.3 \\
            VideoLISA-3.8B (One-Token-Seg-One) & 46.1 & 42.4 & 49.8 \\
            VideoLISA-3.8B (One-Token-Seg-All) & 51.7 & 48.4 & 54.9 \\
            VideoLISA-3.8B (Post optimization) & \textbf{54.5} & \textbf{50.9} & \textbf{58.1} \\
            \specialrule{.1em}{.05em}{.05em}
	\end{tabular} }
	\label{tab:tracking_module_mini}
    \end{minipage}
    \vspace{-8pt}
\end{table}

We conduct ablation studies on various design choices of our model.
The detailed experiment results are provided in the appendix.
Here, we summarize the main takeaways for each study.

\noindent\textbf{Ablation of temporal learning module.}
In this study of Tab.~\ref{tab:abl_temporal_model_mini}, we compare our Sparse Dense Sampling strategy with various design choices, including LISA~\cite{lai2023lisa} finetuned on videos, the most straightforward solution that directly concatenate visual tokens from multiple frames ($n$-frame), the strategy that pools along spatial and temporal dimension separately (ST Pooling), the strategy that pools each frame with different strengths in a slow fast pace.
The comparison of the experiment results shows that our Sparse Dense Sampling strategy outperforms other video-LLM training (sampling) strategies.
In addition to demonstrating the effectiveness of our method, this study also reveals the unique properties of the VOS task.
On the one hand, it requires detailed visual information for accurate segmentation, which makes the pooling-based strategies yield inferior results.
On the other hand, temporal information is also necessary for the model to comprehend motions and behaviors, as validated by the comparison between $n$-frame and ours.

\noindent\textbf{Ablation of temporal association module.}
The main takeaway of this part, as shown in Table~\ref{tab:abl_temporal_model_mini}, lies in the comparison between our method and extensions of image-based LISA.
Specifically, we upgrade LISA to fit the VOS task by 1) (baseline) using one \texttt{<SEG>} token from the first frame to segment subsequent frames, 2) marrying LISA with an off-the-shelf tracking model.
With the help of the tracker, LISA performs clearly better than the baseline, while still performs worse than our method.
The main issue comes from that without perception of the video, the model is incapable of processing queries that are concerned with the full video content and temporal dynamic.
We further quantify the effect of the One-Token-Seg-All approach by contrasting it with a strawman setting, One-Token-Seg-One.
The comparison clearly validates the effect and necessity of the One-Token-Seg-All approach.

\section{Limitation and Future Work}
\label{sec:limitation}
Despite the remarkable performance shown on various benchmarks, our model still has limitations.
We discuss them in this section to inspire future work.
First, our model exhibits deficiencies in computational efficiency.
Although we have already reduced the size of LLM to 3.8B, which is much smaller than previous models (7B, 13B), it still incurs a relatively high computational cost compared to previous work on video object segmentation.
In other words, introducing a MLLM brings remarkable understanding and reasoning ability to the model, while also inducing computational costs.
Exploring methods to achieve a trade-off between these aspects presents an interesting avenue for future research.
Second, we observe that state-of-the-art approaches to video object segmentation often employ dedicated video backbones to enhance performance.
Intuitively, using vision encoder pre-trained on videos would be beneficial for temporal-related tasks, such as object tracking.
However, integrating a video backbone while ensuring compatibility with LLM and SAM decoder is non-trivial.
In this work, we focus on empowering video segmentation tasks with reasoning capabilities based on LLM.
Exploring the integration of a video backbone represents a potential avenue for future research.

\section{Conclusion}
In this work, we propose VideoLISA, a video-based LLM designed for language instructed reasoning segmentation in videos.
It leverages the reasoning capabilities of LLM and employs SAM to produce segmentation masks.
To address the unique challenges in marrying LLM with video object segmentation, we propose two key innovations.
Firstly, a Sparse Dense Sampling strategy is designed to enable LLM to capture and understand temporal dynamics in videos.
By leveraging the inherent temporal redundancy property of videos, this strategy achieves a delicate balance between preserving visual details and temporal context, making it favorable for video object segmentation tasks.
Secondly, we propose a One-Token-Seg-All approach to achieve temporally consistent segmentation masks in the promptable mask decoding paradigm.
Based on a dedicated investigation of the potential and challenges associated with using a single unified prompt to segment video frames, we enhance this capability from both input information foundation and training objective perspectives.
Extensive ablation studies have investigated the function and rationale of the design choices of two modules.
Equipped with the two designs above, our VideoLISA model shows impressive capabilities in video object segmentation, particularly emphasizing complex reasoning, temporal understanding, and object tracking, as validated by our newly organized ReasonVOS benchmark.
Furthermore, it demonstrates notable performance on image segmentation tasks, positioning it as a potential unified model for language-instructed object segmentation.

\clearpage

{
    \small
    \bibliographystyle{plain}
    \bibliography{main}
}

\newpage
\appendix

\section{Appendix}

\subsection{Evaluation on Image Segmentation}
In this section, we evaluate our VideoLISA model on the referring image segmentation task with three widely adopted benchmarks.
The results are presented in Tab.~\ref{tab:refer_seg_img}.
On the refCOCO and refCOCO+ benchmarks, our VideoLISA achieves comparable performance with the image-based LISA model.
On the refCOCOg benchmark, VideoLISA outperforms previous methods, achieving state-of-the-art performance.
In general, the results of this experiment, along with the image reasoning segmentation results shown in the main paper, effectively demonstrate that our VideoLISA model is a strong competitor in image segmentation tasks.

\begin{table*}[h]
    \footnotesize
    \centering
    \caption{Referring segmentation results (cIoU) among ours and existing methods.}
    \vspace{0.1cm}
    \label{tab:refer_seg_img}   
    \tabcolsep=0.3cm
    {
        \begin{tabular}{ l | c c c | c c c | c c }
            \toprule
            
            \multirow{3}*{Method} & \multicolumn{3}{c|}{refCOCO} & \multicolumn{3}{c|}{refCOCO+}  & \multicolumn{2}{c}{refCOCOg} \\ 
            
            \specialrule{0em}{0pt}{1pt}
            \cline{2-9}
            \specialrule{0em}{0pt}{1pt}
            
            ~ & val & testA & testB & val & testA & testB & val(U) & test(U) \\ 
            
            \specialrule{0em}{0pt}{1pt}
            \hline
            \specialrule{0em}{0pt}{1pt}

            MCN~\cite{luo2020multi} & 62.4 & 64.2 & 59.7 & 50.6 & 55.0 & 44.7 & 49.2 & 49.4 \\

            VLT~\cite{ding2021vision} & 67.5 & 70.5 & 65.2 & 56.3 & 61.0 & 50.1 & 55.0 & 57.7 \\

            CRIS~\cite{wang2022cris} & 70.5 & 73.2 & 66.1 & 62.3 & 68.1 & 53.7 & 59.9 & 60.4 \\

            LAVT~\cite{yang2022lavt} & 72.7 & 75.8 & 68.8 & 62.1 & 68.4 & 55.1 & 61.2 & 62.1 \\
            
            ReLA~\cite{liu2023gres} & 73.8 & 76.5 & 70.2 & \textbf{66.0} & \textbf{71.0} & \textbf{57.7} & 65.0 & 66.0 \\
            
            X-Decoder~\cite{zou2023generalized} & - & - & - & - & - & - & 64.6 & -  \\

            SEEM~\cite{zou2023segment} & - & - & - & - & - & - & 65.7 & -    \\

            LISA-7B~\cite{lai2023lisa} & \textbf{74.1} & 76.5 & \textbf{71.1} & 62.4 & 67.4 & 56.5 & 66.4 & 68.5 \\
            
            \specialrule{0em}{0pt}{1pt}
            \hline
            \specialrule{0em}{0pt}{1pt}
            
            \textbf{VideoLISA-3.8B (Ours)} & 73.8 & \textbf{76.6} & 68.8 & 63.4 & 68.8 & 56.2 & \textbf{68.3} & \textbf{68.8} \\

            \bottomrule            
        \end{tabular}
    }
\end{table*}

\subsection{Ablation Studies}

In this section, we present ablation studies on the temporal learning module (the Sparse Dense Sampling strategy), the temporal mask association module (the One-Token-Seg-All approach), and the training data recipe.
For fair comparisons, unless specified, all VideoLISA variants are uniformly trained with the same training setting: 1) 3k iterations in total, 2) the same training data recipe, 3) the same learning rate scheduler, and 4) the same training objective.
Three benchmarks are used for analysis:
1) ReasonSeg~\cite{lai2023lisa} evaluates the reasoning ability of the model;
2) MeViS~\cite{ding2023mevis} reflects the model's performance on temporal learning;
and 3) Ref-DAVIS-17~\cite{refdavis} measures the general RVOS capability of the model.
For evaluation on video benchmarks, the performance metrics of VideoLISA are computed using the simple One-Token-Seg-All approach without post-optimization, revealing the model's essential capabilities.

\subsubsection{Temporal Learning Module}
\label{sec:abl_temporal_learning}
\begin{table*}[t]
	\centering
	\caption{Ablation study on the temporal modeling architecture. *LISA-7B is reproduced using the released codebase.}
	\resizebox{0.98\textwidth}{!}{
		\setlength\tabcolsep{8pt}
		\renewcommand\arraystretch{1.0}
		\begin{tabular}{l|cc|ccc|ccc}
			\specialrule{.1em}{.05em}{.05em}
			\multirow{2}{*}{Method} & \multicolumn{2}{c|}{ReasonSeg (val)} &\multicolumn{3}{c|}{MeViS (valid\_u)} &\multicolumn{3}{c}{Ref-DAVIS-17}  \\
			\cline{2-9}
			& giou & ciou & $\mathcal{J}$\&$\mathcal{F}$ & $\mathcal{J}$ & $\mathcal{F}$  & $\mathcal{J}$\&$\mathcal{F}$ & $\mathcal{J}$ & $\mathcal{F}$ \\ 
            \midrule
            LISA-7B* (Baseline) & 51.7 & 56.7 & 43.2 & 39.9 & 46.5 & 58.8 & 55.1 & 62.5 \\
            LISA-7B* (Vid. FT)  & 48.6 & 56.2 & 44.8 & 41.1 & 48.6 & 58.5 & 54.6 & 62.5 \\
            VideoLISA-3.8B ($n$-frame)  & 55.6 & \textbf{60.8} & 49.9  & 46.7 & 53.0  & 65.5 & 62.2 & 68.9 \\
            VideoLISA-3.8B (Spatial \& Temporal Pooling~\cite{video-chat-gpt})  & 56.0 & 59.9 & 50.8 & 47.8 & 53.8 & 62.2 & 58.4 & 66.3 \\
            VideoLISA-3.8B (Slow-Fast Pooling~\cite{huang2024lita}) & 54.0 & 54.4 & 50.2 & 47.2 & 53.1 & 65.7 & 62.1 & 69.4 \\
            VideoLISA-3.8B (Sparse Dense Sampling) & \textbf{58.9} & 60.0 & \textbf{51.7} & \textbf{48.4} & \textbf{54.9} & \textbf{67.8} & \textbf{64.3} & \textbf{71.3} \\
            
            \specialrule{.1em}{.05em}{.05em}
	\end{tabular} }
	\label{tab:abl_temporal_model}
\end{table*}
In Tab~\ref{tab:abl_temporal_model}, we compare various strategies for temporal learning.
The first row shows the vanilla LISA-7B model, which only focuses on image-based reasoning segmentation.
To infer LISA-7B on video data, we employ a similar One-Token-Seg-All strategy, where the \texttt{<TRK>} token (called \texttt{[SEG]} in the original LISA) comes from the first frame.
This performance serves as a baseline for comparison.
In the second row, we construct a naive solution to adapt LISA to the video domain.
Specifically, we finetune LISA-7B on the aforementioned video segmentation datasets.
The results show that simply finetuning on video data does not significantly improve video performance and even hurts the performance on image reasoning segmentation.
Although training on video datasets may enhance the model's ability to understand temporally related text queries, it still lacks temporal modeling ability from video data, resulting in undesirable performance.

Next, we compare various temporal learning strategies within the VideoLISA framework using the One-Token-Seg-All training objective.
We first experiment with a straightforward video training strategy, called $n$-frame, which directly concatenates the visual features from $n$ sampled frames as input to the large language model.
In our implementation, the value of $n$ is set to the same as $T_{\rm dense}$ for comparison.
As shown in the third row, we observe that with this simple strategy, the model achieves surprisingly good performance across the benchmarks, significantly outperforming LISA-based methods.
Exposure to multiple frames enables the model to perceive temporal dynamics, while the One-Token-Seg-All training objective supervises the model in learning mask association over the temporal dimension, thereby improving multimodal reasoning and temporal consistency in segmentation.
However, due to computational limits, it is prohibitive to include too many frames as it would result in a large number of tokens.

To enable long temporal context perception, we experiment with several pooling strategies, including pooling along the spatial and temporal dimensions separately~\cite{video-chat-gpt}, pooling with different strengths in a slow-fast pace~\cite{feichtenhofer2019slowfast,huang2024lita}, and our Sparse Dense Sampling strategy.
The comparison in Tab.~\ref{tab:abl_temporal_model} reveals that our Sparse Dense Sampling strategy is a more favorable setting among the experiment designs.
The first spatial-temporal pooling strategy eliminates valuable visual details of the video, resulting in inferior performance.
The second slow-fast paced pooling strategy is similar to ours in implementation.
The key difference is that it applies pooling to all frames, albeit with different strengths, while ours preserves the full visual details of the dense frames.
This difference leads to the observed performance gap.
We argue that this difference is significant due to the unique nature of the video object segmentation task.
On one hand, it requires detailed visual information for accurate segmentation, causing pooling-based strategies to yield inferior results.
On the other hand, the temporal dimension is also necessary for the model to comprehend motions and behaviors, as validated by the comparison between the $n$-frame approach and ours.
Although recent studies~\cite{xu2024pllava} show that applying pooling to visual tokens does not affect the performance of VQA tasks, our experiments validate that preserving the full resolution of visual tokens is necessary for dense prediction tasks, and applying pooling leads to sub-optimal results.

\begin{table*}[t]
	\centering
	\caption{Ablation study on the mask association \textit{i.e.,} tracking architecture. *LISA-7B is reproduced using the released codebase.}
	\resizebox{0.98\textwidth}{!}{
		\setlength\tabcolsep{8pt}
		\renewcommand\arraystretch{1.0}
		\begin{tabular}{l|ccc|ccc}
			\specialrule{.1em}{.05em}{.05em}
			\multirow{2}{*}{Method} &\multicolumn{3}{c|}{MeViS (valid\_u)} &\multicolumn{3}{c}{Ref-DAVIS-17}  \\
			\cline{2-7}
			& $\mathcal{J}$\&$\mathcal{F}$ & $\mathcal{J}$ & $\mathcal{F}$  & $\mathcal{J}$\&$\mathcal{F}$ & $\mathcal{J}$ & $\mathcal{F}$ \\ 
            \midrule
            LISA-7B* (Baseline) & 43.2 & 39.9 & 46.5 & 58.8 & 55.1 & 62.5 \\
            LISA-7B* + XMem\cite{cheng2022xmem} & 45.6 & 41.9 & 49.3 & 62.7 & 60.0 & 65.5 \\
            VideoLISA-3.8B (One-Token-Seg-One) & 46.1 & 42.4 & 49.8 & 60.2 & 56.5 & 63.8 \\
            VideoLISA-3.8B (One-Token-Seg-All) & 51.7 & 48.4 & 54.9 & 67.8 & 64.3 & 71.3 \\
            VideoLISA-3.8B (Post optimization) & \textbf{54.5} & \textbf{50.9} & \textbf{58.1} & \textbf{68.7} & \textbf{65.5} & \textbf{72.0} \\
            
            \specialrule{.1em}{.05em}{.05em}
	\end{tabular} }
	\label{tab:tracking_module}
\end{table*}

\subsubsection{Temporal Association Module}

In Tab.~\ref{tab:tracking_module}, we compare the design choices for the temporal association module, \textit{i.e.,} tracking.
As in previous comparisons, the One-Token-Seg-All strategy in LISA-7B serves as the baseline in the first row.
One straightforward solution based on LISA is to plug an off-the-shelf tracker into the model.
During inference, LISA outputs the segmentation mask of the first frame based on language instruction.
The tracker then tracks the segmented object through the video, yielding segmentation masks for the subsequent frames.
Specifically, we adopt the popular XMem~\cite{cheng2022xmem} model as the tracker, as shown in the second row of the table.
Compared to VideoLISA (both One-Token-Seg-All and post-optimization), LISA+XMem achieves worse performance on these benchmarks.
This validates that simply plugging an existing tracker into an image-based reasoning segmentation model does not address the problem of video reasoning segmentation.
The vital issue is that the LLM in charge of perception and reasoning does not capture the entire video content, making its predictions nonsensical.
In contrast, VideoLISA's temporal learning module and dedicated training objective enrich the \texttt{<TRK>} token with semantic information, enabling it to find the target object across all frames.

To quantify the effect of the One-Token-Seg-All training objective, we build a strawman setting named One-Token-Seg-One.
In this setting, the video content is captured with the temporal learning module, but the training only supervises the segmentation of one frame.
The comparison is shown in the third and fourth rows of Tab.~\ref{tab:tracking_module}.
We observe that the slight difference in supervision leads to a significant performance gap in the benchmarks.
This indicates that the One-Token-Seg-All training objective is essential for achieving temporally consistent masks.

In the last row, we present post-optimization, which leverages both the reasoning and segmentation abilities of VideoLISA and a mature tracking model.
Specifically, we first use VideoLISA to produce the \texttt{<TRK>} token and then use it to segment the sampled dense frames.
Then, the post-optimization model, implemented as XMem++~\cite{bekuzarov2023xmem++}, takes dense frames and their segmentation masks as references in its permanent memory and infers the masks for the remaining frames.
The reasons for choosing the dense frames as the mask reference include: 1) the dense frames are seen by VideoLISA, thus their masks should be more accurate than those of other unseen frames, and 2) the dense frames are intentionally sampled from the video in a uniform manner, naturally providing a long-range yet diverse reference signal.
By leveraging the association ability from the post-optimization step, VideoLISA achieves the best performance.

\subsubsection{Ablation on Training Data}
\begin{table*}[t]
	\centering
	\caption{Ablation study on the training data recipe.}
	\resizebox{0.98\textwidth}{!}{
		\setlength\tabcolsep{8pt}
		\renewcommand\arraystretch{1.0}
		\begin{tabular}{cccc|cc|ccc|ccc}
			\specialrule{.1em}{.05em}{.05em}
			\multicolumn{4}{c|}{Training Data} & \multicolumn{2}{c|}{ReasonSeg (val)} &\multicolumn{3}{c|}{MeViS (valid\_u)} &\multicolumn{3}{c}{Ref-DAVIS-17}  \\
            \hline
			Image Seg. & Video Seg. & Image QA & Video QA & giou & ciou & $\mathcal{J}$\&$\mathcal{F}$ & $\mathcal{J}$ & $\mathcal{F}$  & $\mathcal{J}$\&$\mathcal{F}$ & $\mathcal{J}$ & $\mathcal{F}$ \\ 
            \midrule
            \cmark &        &        &       & 57.2 & 60.0 & 46.0 & 43.3 & 48.6 & 62.6 & 58.9 & 66.3 \\
                   & \cmark &        &       & 41.4 & 46.5 & 49.3 & 45.8 & 52.8 & 66.0 & 62.7 & 69.3 \\
            \cmark & \cmark &        &       & 58.9 & 60.0 & 51.7 & 48.4 & 54.9 & \textbf{67.8} & \textbf{64.3} & \textbf{71.3} \\
            \cmark & \cmark & \cmark &       & 56.0 & 65.6 & 49.8 & 46.8 & 52.9 & 66.8 & 63.4 & 70.3 \\
            \cmark & \cmark & \cmark &\cmark & \textbf{60.6} & \textbf{67.4} & \textbf{52.0} & \textbf{49.3} & \textbf{54.8} & 66.9 & 63.5 & 70.3 \\
            
            \specialrule{.1em}{.05em}{.05em}
	\end{tabular} }
	\label{tab:abl_data}
\end{table*}

Our model undergoes joint training on both image and video datasets.
An investigation of the training data is presented in Tab.~\ref{tab:abl_data}.
We first observe that with image-only segmentation datasets, the model achieves decent performance in reasoning segmentation.
However, the performance on video benchmarks is unsatisfactory, possibly due to insufficient temporal information in the training data.
When using video-only segmentation settings, compared to image-only, the performance on video benchmarks increases significantly.
Simultaneously, the model experiences a dramatic drop in performance in reasoning segmentation.
This comparison demonstrates that video training is helpful for the VOS task, while image data is also necessary to exploit the reasoning ability of the model.
When combining the image and video segmentation datasets, the model yields remarkable performance across various benchmarks.

Next, we additionally explore the effect of using visual question answering (VQA) data.
We first observe that after adding Image-QA data into training, the model experiences a slight performance drop in all benchmarks.
Then, with the involvement of Video-QA data, the model achieves much better performance on the reasoning segmentation benchmark.
Among the two video benchmarks, compared to the model trained with segmentation-only data, this model shows slightly better performance on the MeViS offline validation set yet worse performance on Ref-DAVIS-17.
Intuitively, VQA data has the potential to enhance the model's reasoning ability.
However, it may also make the multi-task training more challenging, as revealed by the performance fluctuation among different benchmarks. Maintaining the compatibility of different types of training data and tasks is left for future work.

\subsection{Qualitative Results}
\label{sec:more_qualitative}
\begin{figure*}[]
    \centering
    \vspace{-10pt}
    \includegraphics[width=0.98\linewidth]{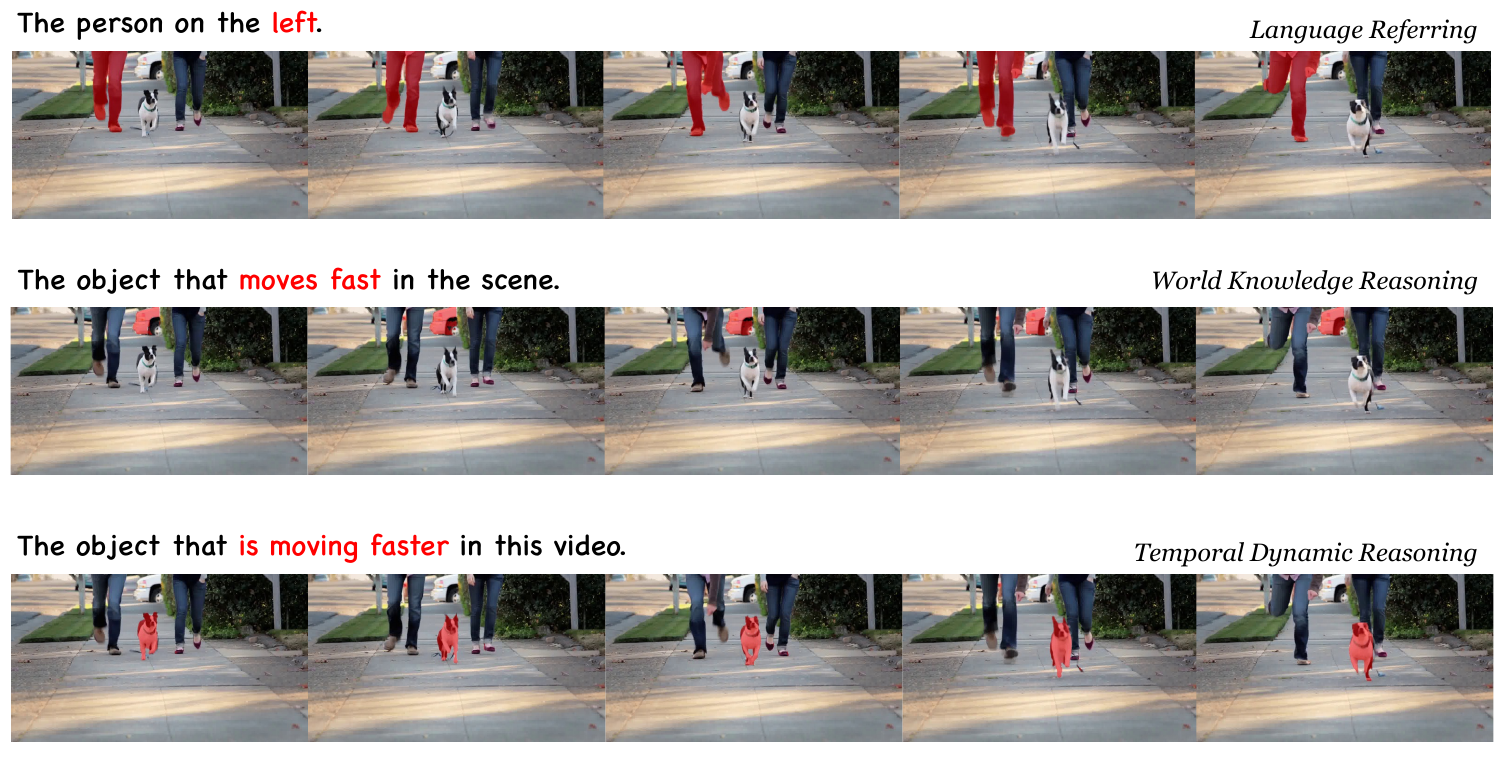}
    \vspace{-8pt}
    \caption{
    VideoLISA is a capable model on video object segmentation with versatile language-instructed reasoning abilities.
    Beyond basic language referring, it enables complex reasoning by leveraging world knowledge and videos temporal dynamics.
    }
    \label{fig:teaser}
\end{figure*}

In Fig.~\ref{fig:teaser}, we use a representative video to showcase the versatile language-instructed reasoning capabilities of our model.
VideoLISA can do segmentation in videos via language referring, world knowledge reasoning, and video temporal reasoning.
Additionally, the model can discern subtle differences in language instructions and is not biased to salient or moving objects.

\begin{figure*}[]
    \centering
    \includegraphics[width=0.98\linewidth]{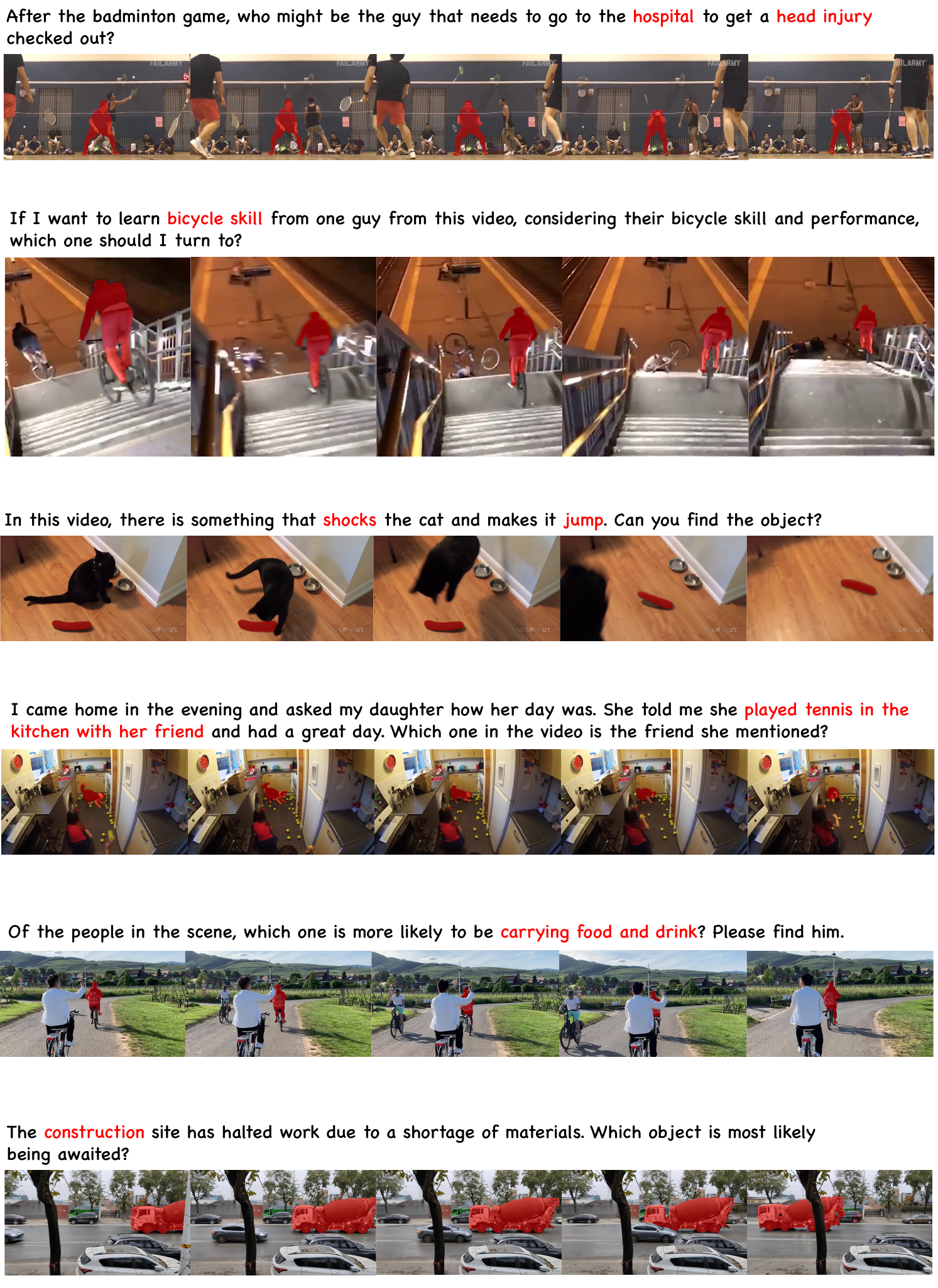}
    \caption{
    More qualitative examples of VideoLISA.
    }
    \label{fig:qual_1}
\end{figure*}

\begin{figure*}[]
    \centering
    \includegraphics[width=0.98\linewidth]{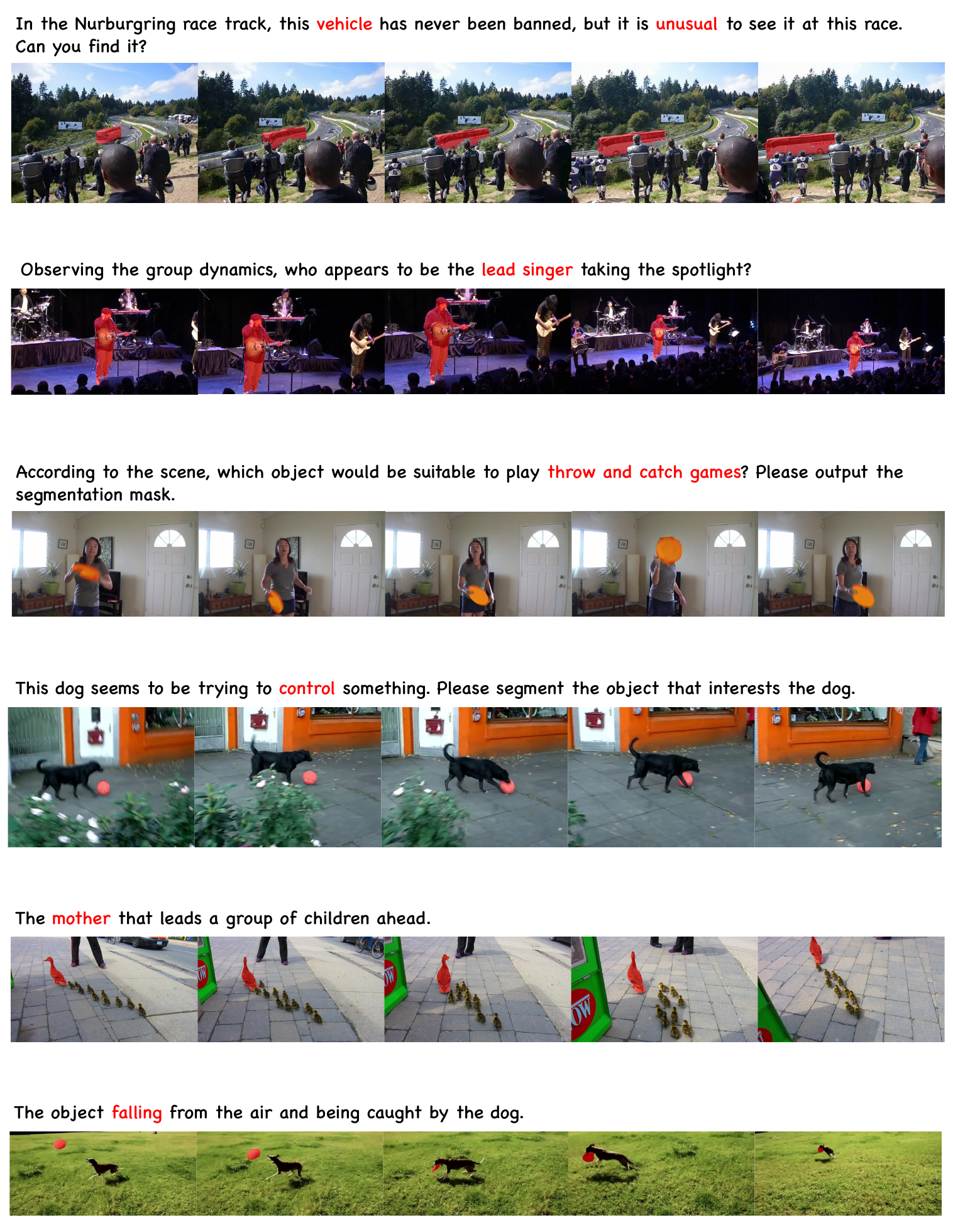}
    \caption{
    More qualitative examples of VideoLISA.
    }
    \label{fig:qual_2}
\end{figure*}

In Fig.~\ref{fig:qual_1} and Fig.~\ref{fig:qual_2}, we provide more abundant qualitative examples of VideoLISA.
The red text is only for illustration purposes.
No special prompting techniques were employed.
It's important to note that these examples were generated using the One-Token-Seg-All inference approach without post-optimization.

\begin{figure*}[]
    \centering
    \includegraphics[width=0.98\linewidth]{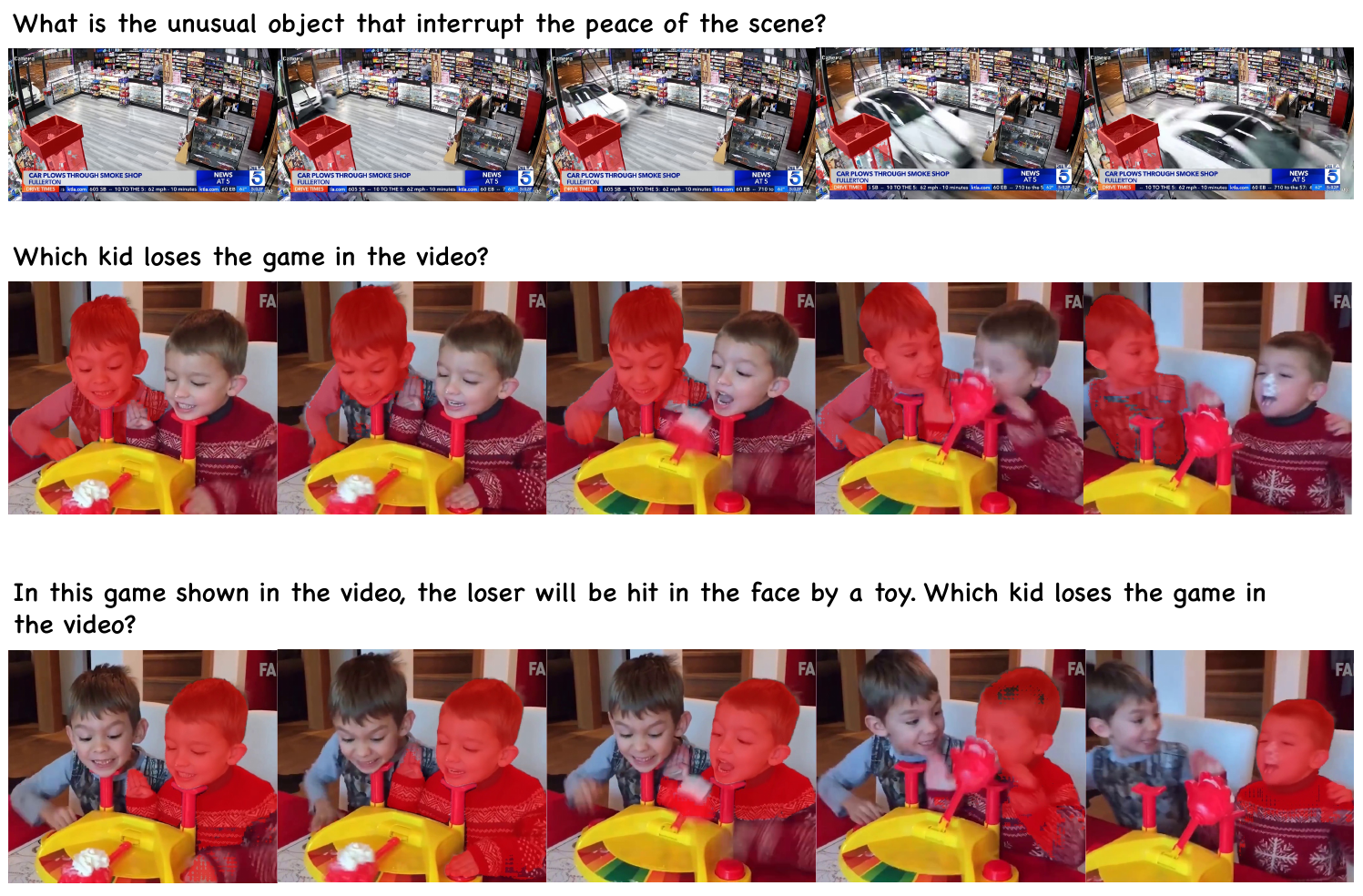}
    \caption{
    Failure cases of VideoLISA.
    }
    \label{fig:fail}
\end{figure*}

\subsection{Failure Cases}
\label{sec:failure}
To understand the limitations and capability boundaries of our method, we analyze several failure cases as shown in Fig.~\ref{fig:fail}.
In the first example, the video shows a car crashing into a grocery store.
We prompt the model to find the unusual object that interrupts the peace of the scene.
Although we try to rephrase the prompt in various ways, the model consistently outputs the object in the bottom left corner.
We hypothesize that the issue stems from the inherent hallucination of the MLLM, which recognizes the object as a stove, a telephone pole, or something else.

In the second example, we ask the model to find the kid who loses the game
We humans have the background knowledge to determine the match result.
However, it seems like this game is beyond the knowledge scope of the MLLM, causing it to segment the wrong person.
Consequently, we provide some background information about the game rules in the text prompt and then ask the same question.
As shown in the third example of Fig.~\ref{fig:fail}, with this cue, the model is able to segment the correct person.
These examples demonstrate that the reasoning capabilities of VideoLISA are bounded by the multimodal large language model behind it, yet this can be alleviated by prompt engineering techniques.
The third example also exhibits low-quality segmentation masks in certain frames, leaving room for future improvements.

\begin{figure*}[]
    \centering
    \includegraphics[width=0.75\linewidth]{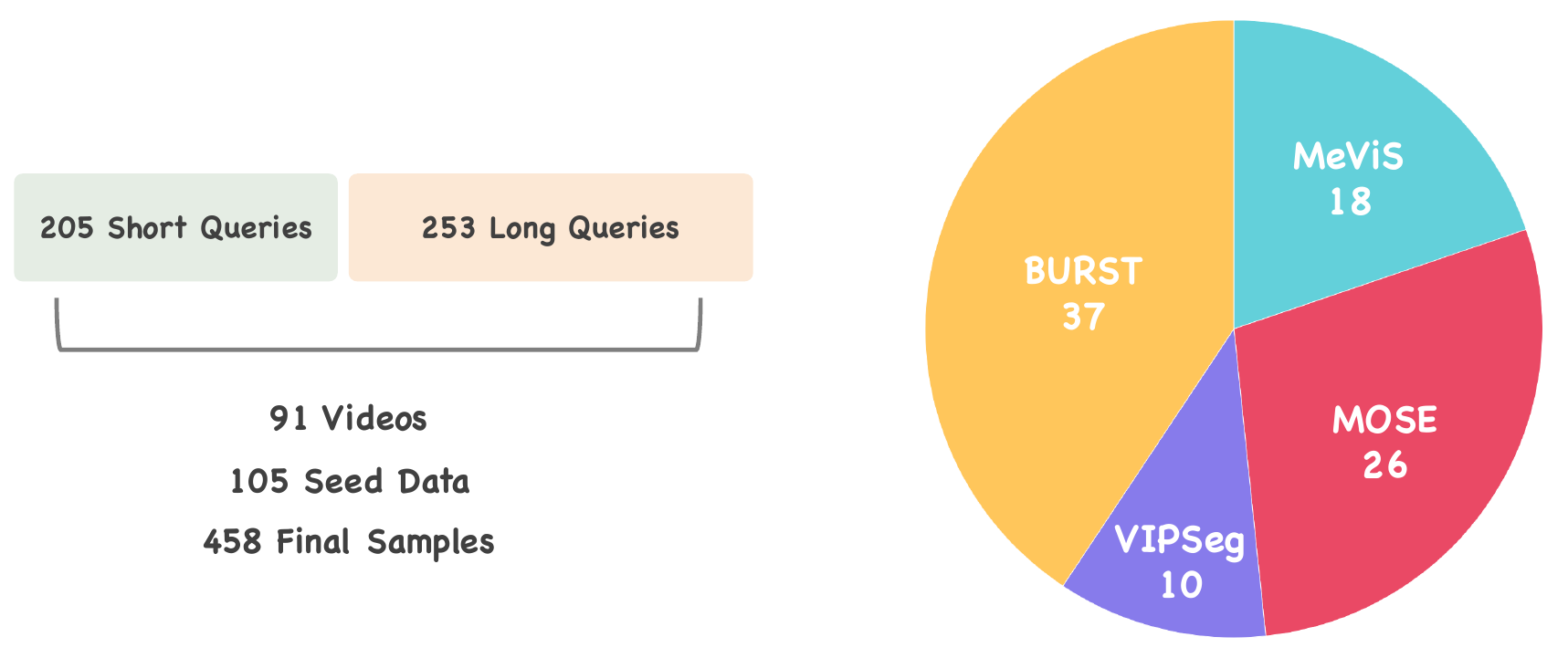}
    \caption{
    ReasonVOS benchmark. The left part shows the statistics of data samples. The right part shows the source of the videos.
    }
    \label{fig:bench}
\end{figure*}

\vspace{-5pt}
\section{Benchmark}
\vspace{-5pt}
We show the data statistics of our ReasonVOS benchmark in Fig.~\ref{fig:bench}.
We select videos and mask annotations from various sources and annotate additional text descriptions.
In total, ReasonVOS consists of 91 videos.
We manually annotate 105 video-instruction-mask samples as seed data and use Claude 3 API to augment the data into 458 samples.
We further categorize the text descriptions into short query and long query.
Short queries are descriptions of specific objects, usually in the format of attributive clauses.
Long queries are instructions that require reasoning, usually in the format of a full sentence.

\section{Broader Impact}
\label{sec:broader_impact}
The development of our reasoning-based video segmentation model holds significant potential for transforming a variety of fields by enhancing the ability to analyze and interpret video content.
In the realm of surveillance, this technology can improve security measures by accurately identifying and tracking suspicious behavior, thereby preventing potential threats.
In educational settings, the model can assist teachers in identifying and addressing student engagement patterns, fostering a more responsive learning environment.
For healthcare, our model can be applied to monitor patient activities, supporting early intervention and personalized care strategies.
Additionally, in everyday scenarios, such as pet care or home organization, this technology can assist individuals in making informed decisions quickly and efficiently.
By leveraging advanced reasoning capabilities, our model not only advances the field of computer vision but also provides practical solutions that enhance safety, learning, health, and daily life.
However, it is crucial to consider ethical implications, such as privacy concerns and the potential for misuse, ensuring that these technologies are implemented responsibly and equitably.

\clearpage

\end{document}